\documentclass[sigconf, authorversion, nonacm]{acmart}
\renewcommand\footnotetextcopyrightpermission[1]{} 
\usepackage{url}
\usepackage{latexsym}
\usepackage{graphicx}
\usepackage{algorithm}
\usepackage{algpseudocode}
\usepackage{caption}
\usepackage{array}
\usepackage{makecell}
\usepackage{amsmath}
\usepackage{adjustbox} 

\usepackage{enumitem}
\usepackage[utf8]{inputenc}

\AtBeginDocument{%
  \providecommand\BibTeX{{%
    \normalfont B\kern-0.5em{\scshape i\kern-0.25em b}\kern-0.8em\TeX}}}

\author{Anand A. Rajasekar}
\authornote{Both authors contributed equally to this research.}
\author{Praveen Tangarajan*}
\author{Anjali Nainani}
\author{Amogh Batwal}
\author{Vinay Rao Dandin}
\author{Anusua Trivedi}
\author{Ozan Ersoy}
\affiliation{%
  \institution{Flipkart US R\&D Center}
  \country{Bellevue, Washington, USA}
}

\renewcommand{\shortauthors}{Anand and Praveen, et al.}

\usepackage{float}

\title{Flippi: End To End GenAI Assistant for E-Commerce}
\begin{document}

\begin{abstract}
The emergence of conversational assistants has fundamentally reshaped user interactions with digital platforms. This paper introduces Flippi-a cutting-edge, end-to-end conversational assistant powered by large language models (LLMs) and tailored for the e-commerce sector. Flippi addresses the challenges posed by the vast and often overwhelming product landscape, enabling customers to discover products more efficiently through natural language dialogue. By accommodating both objective and subjective user requirements, Flippi delivers a personalized shopping experience that surpasses traditional search methods. This paper details how Flippi interprets customer queries to provide precise product information, leveraging advanced NLP techniques such as Query Reformulation, Intent Detection, Retrieval-Augmented Generation (RAG), Named Entity Recognition (NER), and Context Reduction. Flippi’s unique capability to identify and present the most attractive offers on an e-commerce site is also explored, demonstrating how it empowers users to make cost-effective decisions. Additionally, the paper discusses Flippi’s comparative analysis features, which help users make informed choices by contrasting product features, prices, and other relevant attributes. The system’s robust architecture is outlined, emphasizing its adaptability for integration across various e-commerce platforms and the technological choices underpinning its performance and accuracy. Finally, a comprehensive evaluation framework is presented, covering performance metrics, user satisfaction, and the impact on customer engagement and conversion rates. By bridging the convenience of online shopping with the personalized assistance traditionally found in physical stores, Flippi sets a new standard for customer satisfaction and engagement in the digital marketplace.
\end{abstract}
\maketitle

\section{Introduction}
Conversational commerce has rapidly evolved, enabling consumers to shop, purchase, and seek assistance through interactive platforms such as live chat, voice assistants, and messaging apps. These conversational tools are pivotal in e-commerce, guiding customers to discover suitable products, make informed purchasing decisions, and access post-purchase support.

Historically, online shopping has been dominated by search bar-driven navigation, where users begin with broad queries-such as product type and basic requirements like brand or features. However, traditional platforms often present significant challenges that detract from the shopping experience. Chief among these is choice overload: the sheer volume of available products can make it difficult for users to find what they want. Compounding this are search algorithms that sometimes misinterpret queries, leading to irrelevant results. Filtering and sorting tools may lack the necessary granularity or intuitiveness, making it hard for users to efficiently narrow their options.

A further challenge is the lack of personalized recommendations, which are increasingly essential for aligning product suggestions with individual tastes and preferences. Static product displays fail to deliver the interactive, engaging experiences that users now expect, often resulting in decision fatigue. Product pages are typically dense with information-ranging from seller-provided descriptions to customer reviews-making it difficult for users to extract relevant details, especially on mobile devices with limited screen space. For value-driven customers seeking the best offers, the decision-making process becomes even more complex. 
Thus, developing a comprehensive conversational assistant that supports users throughout their e-commerce journey-from product discovery to customer support-is both necessary and impactful.However, building such a system presents unique challenges. Multi-turn conversations require the assistant to understand and retain the context of previous interactions, ensuring a natural conversational flow and preventing repetitive exchanges. The ability to ask targeted follow-up questions is also crucial for a complete conversational experience.

Technical integration with existing e-commerce systems is another hurdle. The architecture must be streamlined to interact seamlessly with various components, such as search engines, product detail fetchers, and customer support and experience (CX) systems. The complexity of integrating with diverse engineering elements can be substantial.

This paper presents an end-to-end data science architecture for modeling e-commerce assistants. We detail the current architecture, the design decisions behind it, and the individual modules and their evaluations. The paper is organized as follows: Section 2 reviews related work; Section 3 outlines the proposed framework; Section 4 discusses the evaluation methodology; Section 5 presents results; Section 6 examines Flippi’s production impact; and Section 7 concludes with future directions

\section{Related Work}
The rapid advancement of natural language processing (NLP) and machine learning has driven remarkable growth in conversational agents and chatbots, particularly within e-commerce~\cite{lim2022alexa,lim2022conversational}. 
Early e-commerce applications leveraged conversational interfaces to improve shopping experiences, such as chatbot-based recommenders that utilized user preferences and browsing history for personalized suggestions~\cite{liu2019deep}, and dialogue generation models that enabled more intuitive product discovery~\cite{ZHANG2022185}. Further studies have explored the influence of chatbot language style on user trust~\cite{cheng2022chatbot}, negotiation capabilities in conversational commerce~\cite{priyanka2023product}, and factors affecting customer engagement among younger demographics~\cite{rumagit2023chatbots}.

The emergence of Large Language Models (LLMs) has marked a paradigm shift in NLP, enabling significant progress in knowledge acquisition and generation~\cite{naveed2024comprehensive,zhao2023survey,minaee2024large}. Trained on extensive corpora using self-supervised objectives~\cite{gui2023survey}, models such as BERT~\cite{DBLP:journals/corr/abs-1810-04805}, GPT-3.5~\cite{DBLP:journals/corr/abs-2110-08207,DBLP:journals/corr/abs-2005-14165}, GPT-4~\cite{openai2024gpt4}, LLaMA~\cite{touvron2023llama}, and Mistral~\cite{jiang2023mistral} have advanced the generation of human-like text and deepened linguistic understanding.

Despite their strengths, adapting LLMs to specific domains and user preferences remains challenging~\cite{kirk2024understanding,havrilla2024teaching}. Approaches such as domain-specific fine-tuning~\cite{hu2021lora,xu2023parameterefficient} and context-aware prompting~\cite{sahoo2024systematic,liu2021pretrain} have been proposed, though fine-tuning can be resource-intensive. Retrieval Augmented Generation (RAG) offers a scalable alternative, augmenting LLMs with external knowledge to improve factuality and reduce hallucinations~\cite{DBLP:journals/corr/abs-2005-11401,gao2024retrievalaugmented,siriwardhana2022improving}. Advanced prompt engineering techniques, further enhances LLM reasoning within RAG frameworks~\cite{wei2023chainofthought,chu2023survey}.

Evaluating LLM-driven systems requires more than traditional NLP metrics, prompting the development of frameworks that assess qualitative aspects such as bias, toxicity, factual consistency, and hallucination rates~\cite{guo2023evaluating,chang2023survey,ribeiro-etal-2020-beyond,zheng2023does}. Human-in-the-loop evaluation remains essential for judging the quality and utility of responses in complex interactive systems~\cite{DBLP:journals/corr/abs-2005-11401,DBLP:journals/corr/abs-2104-14337,schiller2024human}.

Recent adoption of generative AI has led to the development of advanced e-commerce assistants, incorporating features like conversational product discovery, personalized recommendations, product comparisons, and support for order-related queries. These systems leverage large product catalogs, customer reviews, and community Q\&A to create interactive shopping experiences. However, challenges persist in managing hallucinations, maintaining context over multi-turn dialogues, and leveraging structured product data for accurate responses. Achieving deep personalization and robust long-term context retention remains difficult for many solutions.

Building on these foundations, this paper introduces Flippi, an end-to-end generative AI assistant for e-commerce. Flippi integrates advanced NLP techniques—Query Reformulation, Intent Detection, RAG, Named Entity Recognition, and Context Reduction—within a modular, scalable architecture. By addressing key challenges in product discovery, personalization, offer identification, and product comparison, Flippi provides a comprehensive framework for reliable, user-centric conversational assistance in digital marketplaces.

\section{Proposed Architecture}
The interaction between the customer and the assistant begins when the customer inputs their question into the assistant's chat window, thereby starting a session. To facilitate a continuous and coherent conversation, each user query, along with the complete session context (previous turns of dialogue), is forwarded to the Standalone Query (SAQ) module. The primary function of the SAQ module is to leverage the full session history and the current user query to generate a contextually independent query that accurately reflects the customer's complete and most recent intent. This reformulated, standalone query is subsequently directed to a coarse intent classification model, which determines the appropriate processing pipeline, or "flow," for the query. The Flippi framework is designed around several core flows, primarily including Product Search, Decision Assistance (for product-specific Q\&A), Offers (for promotion-related inquiries), and Customer Experience (CX, for post-purchase and customer service issues), with the architecture allowing for scalability to additional use cases.

\begin{figure*}[!htbp]  
    \centering  
    \includegraphics[width=\linewidth]{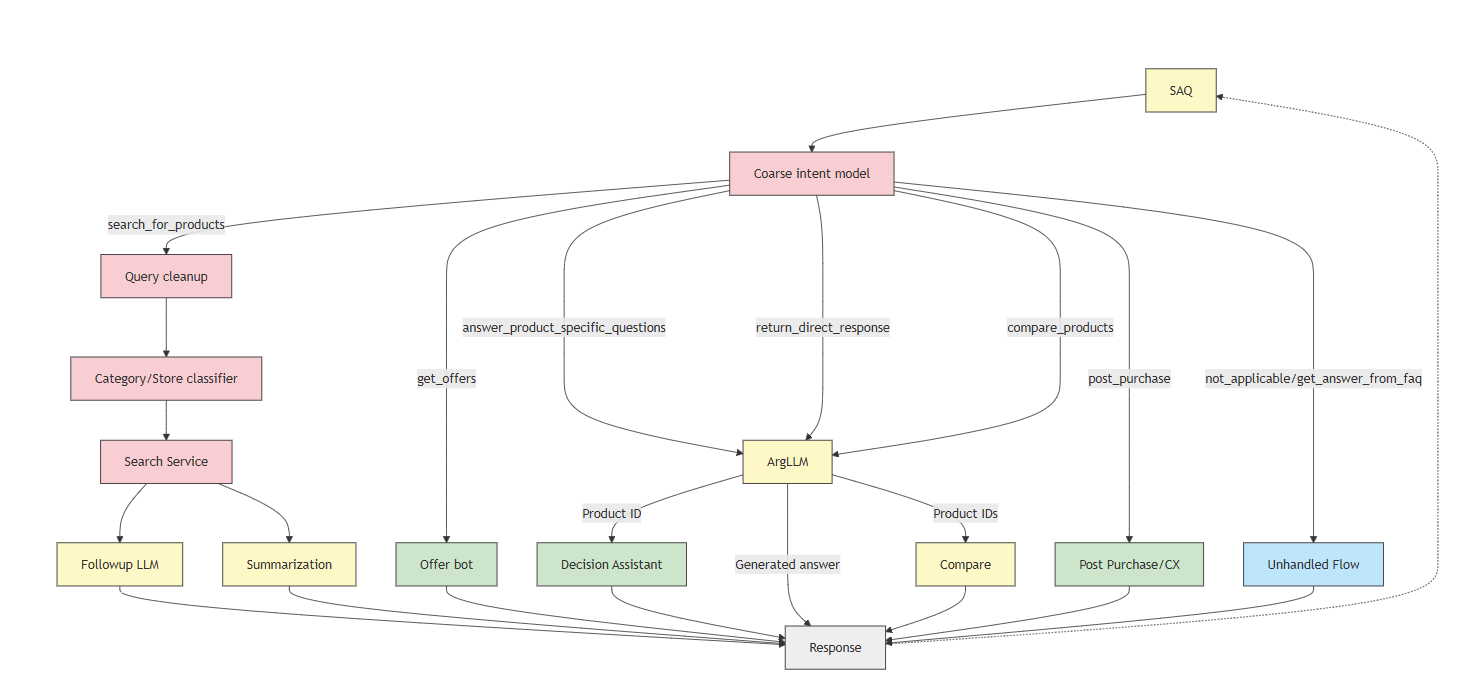}  
    \caption{High level architecture}  
    \label{fig:main-arch}  
\end{figure*}

If the coarse intent model identifies the user's intent as requiring a direct response (e.g., handling chitchat, general e-commerce platform FAQs), decision assistance, or product comparison, the ArgsLLM module is activated. The main functionality of this module is to gather all necessary arguments required for the Decision Assistance and Product Comparison flows. Additionally, it is capable of directly generating a response to the user's query for certain intents. The overall process is illustrated in Figure 1. At a high level: the Search flow aims to retrieve relevant product sets based on customer interest. The Decision Assistance (DA) flow answers product-related questions for a specific item using catalog information and user-generated content (UGC) like reviews. The Offers flow addresses queries about current or upcoming promotions and discounts. The CX flow manages customer service inquiries related to post-purchase events. Finally, other ancillary flows provide templated responses for queries outside the e-commerce scope or answer general FAQs pertaining to the e-commerce platform.
\begin{table*}[h]
    \begin{tabular}{|l |p{7.5 cm} | p{7.5 cm}|}
    \cline{2-3}
\multicolumn{1}{c|}{}& Sample 1 & Sample 2 \\
\hline
Conversation & \makecell[l]{\textbf{User:} oppo mobile\\
\textbf{Bot:} Products that match your query are \\
OPPO A78 5G (Glowing Black),\; OPPO A78 (Aqua Green)\\ and OPPO A38 (Glowing Gold)\\
\textbf{Bot:} What budget do you have in mind?\\
\textbf{User:} 15,000\\
\textbf{Bot:} Products that match your query are\\
OPPO A38 (Glowing Gold),\;OPPO A17 (Lake Blue) \\and OPPO A58 (Dazzling Green)\\
\textbf{Bot:} Any particular feature you are looking for in a phone?\\
\textbf{User:} Motorola mobile}& \makecell[l]{
\textbf{User:} PETER ENGLAND suit for wedding under 6000\\
\textbf{Bot:} Products that match your query are \\
PETER ENGLAND SLIM FIT Solid Suit,\; Raymond Men\\ Solid Suit and VAN HEUSEN Men 2 PC Suit Solid Suit]\\
\textbf{Bot:} Do you have a specific size in mind?\\
\textbf{User:} No, what is the color of VAN HEUSEN suit?}\\
\hline
SAQ & Motorola mobile within 15,000 & What is the color of VAN HEUSEN Men 2 PC Suit Solid Suit?\\
\hline
    \end{tabular}
\caption{Example user conversations with standalone queries}
\label{tab:d9}
\end{table*}
\subsection{Standalone Query Module (SAQ)}
The SAQ module is the first and one of the most crucial components of the Flippi pipeline. It facilitates continuous conversation between user and assistant. The SAQ module receives each user's query along with the complete session context. The session context includes all the previous queries and responses exchanged between the user and the assistant. The primary function of the SAQ module is to generate an independent query that accurately reflects the user's current intent while incorporating the important and relevant information from the context. This revised query should encapsulate all the necessary information, allowing the conversational agent to provide a more targeted and relevant response.

Another function of this component is to perform product disambiguation. Product disambiguation is the process of identifying the correct product that the user is referring to when there are multiple products in the context. The SAQ module uses the context of the conversation to disambiguate the user's request and add the correct product name to the reformulated query. E.g., if the user says "second product", the SAQ module will identify the right product from the bot output and add the product name to the final query. Table \ref{tab:d9} consists of example conversations and the corresponding standalone queries. Currently, this component is powered by an LLM call.

\subsection{Coarse Intent Model}
The primary functionality of this model is to understand the intent behind the query and trigger the appropriate flow to handle the user request. Currently, this model takes the given SAQ query as input and predicts one of the eight distinct intents which then is leveraged as a routing mechanism for downstream flows. 

Here are the following eight intents and their definitions:
\begin{enumerate}[topsep=0pt,itemsep=-1ex,partopsep=1ex,parsep=1ex]
\item \textbf{\textit{answer\_product\_specific\_questions}}: This intent pertains to queries that revolve around detailed information about a product such as its specifications, warranty, payments and availability. The Decision Assistant flow accomplishes this task by utilizing data from both the product catalog and user-generated content, such as customer reviews. For instance, any question about a product's features or functionality will be addressed through this intent.
\item \textbf{\textit{search\_for\_products}}: This intent caters queries aiming to find or explore products on the e-commerce platform. By calling upon a search API, the intent can further clarify the user's needs by posing follow-up questions to better understand their requirements.
\item \textbf{\textit{compare\_products}}: This intent is designed to handle queries that involve comparing two or more products or their features. For example, a query like “RAM of iPhone 14 vs Samsung S23” would be directed to this intent, enabling the user to make a more informed decision about the product they wish to purchase.
\item \textbf {\textit{return\_direct\_response}}: This intent deals with queries that are either greetings, casual conversations or questions seeking general information about product features. These queries may include questions such as “What is an OLED display”, “What is the difference between a processor and RAM” or “What is an operating system”. The main focus of these queries is to provide general information related to shopping.
\item \textbf {\textit{answer\_offer\_related\_questions}}:  This intent handles queries concerning current or upcoming offers. This includes general offers, big sale events, and promotions on specific products. Queries related to discounts and other promotional activities are directed here.
\item \textbf {\textit{post\_purchase}}:  This intent handles customer queries that occur after a purchase has been completed. These may include issues with orders, tracking delivery, refund, or cancellation of a product, and other post-purchase concerns. Queries of this nature are typically directed to the customer service department
\item \textbf{\textit{get\_answer\_from\_faq}}: This intent addresses commonly asked, non-product specific questions that are specific to ecommerce platform. These may include generic questions about payment methods, return policies, exchange policies, or pre-purchase queries. Examples include “How to order?”, “Where can I find the refund policy” or “How to apply an offer code?”.
\item \textbf {\textit{not\_applicable}}: This category includes any queries that fall outside the scope of online shopping or do not fit within the categories mentioned above. Queries that are not related to general conversation, shopping, or product-specific queries are categorized as not applicable. Examples may include "how to calculate the area of a circle” etc.
\end{enumerate}
Different BERT based architectures were tried for this text classification and we were able to achieve high classification accuracy of around ~95\%. 
Given high traffic requirements, we preferred the smaller and low latency DistillBERT model. Data augmentation and MixUp approaches were tried to further improve the model. This model was improved by data augmentation techniques such as continuous sampling using active learning and relabeling by LLMs. Results are discussed in Section 4.

\subsection{Product Search and Followup}
Product search and discovery are core use cases for Flippi. This flow incorporates multiple components and leverages the e-commerce platform's existing search infrastructure. A store classifier, fine-tuned on production queries, is used to identify the relevant product category or store. This flow also includes enhancements beyond standard search to better interpret subjective queries and handle queries that yield no initial results.
First, the SAQ output undergoes a 'Query Cleanup' step to normalize it for the platform's search service. The cleaned query is then passed to the search service, which returns a list of products (e.g., up to 24 products, with a subset, such as 8, displayed initially with a 'View More' option). 
As customers are presented with products pertaining to their initial query, an accompanying question related to the product is posed to gain a more comprehensive understanding of their preferences.

Generating effective follow-up questions for the assistant adheres to several key principles:

\begin{enumerate}[topsep=0pt, itemsep=-0.5ex, partopsep=1ex, parsep=1ex]
    \item \textbf{Relevance to the initial query and current context:} Follow-up questions and their suggested values should be directly pertinent to the user's stated needs and the ongoing conversation.
    \textit{Example: If a user queries "running shoes for wide feet under \$80", follow-up questions should focus on relevant attributes like "preferred cushioning level" or "trail vs. road running", rather than suggesting high-end racing flats or features common in shoes well over the specified budget.}

    \item \textbf{Grounding in available product assortment:} Proposed values or features in follow-up questions must correspond to actual products or characteristics available on the e-commerce platform.
    \textit{Example: When discussing "cotton t-shirts", if the platform primarily stocks common colors, the follow-up should not suggest "artisanal hand-dyed indigo" as a color option if such products are not offered.}

    \item \textbf{Avoidance of repetition:} The same follow-up question or attribute type should generally not be repeated within the same conversational thread for a given search, especially if the user has already declined or ignored a similar prompt.
    \textit{Example: If the assistant previously asked "Are you looking for a specific brand?" and the user responded "No, show me all options", it should avoid asking about brands again immediately unless the search context significantly changes.}

    \item \textbf{Adaptability to context shifts:} The assistant must recognize and adapt to changes in the user's focus or product interest within a single session.
    \textit{Example: If a user initially searches for "digital cameras" and later asks "what about tripods for these models?", subsequent follow-up questions should shift to tripod-specific attributes (e.g., "material preference for the tripod?", "maximum height needed?") rather than persisting with camera features.}

    \item \textbf{Conciseness in interaction:} The number of follow-up questions posed before providing or refining results should be limited to prevent user fatigue and streamline the discovery process.
    \textit{Example: After an initial set of products is displayed, posing more than two or three clarifying questions before showing updated results is generally advisable to maintain engagement.}
\end{enumerate}

For generating these follow-up questions, a context-aware prompt was developed for an LLM to autonomously suggest relevant attributes. This prompt aims to generate pertinent follow-up questions with appropriate value suggestions, while avoiding repetition. However, grounding these suggestions to the e-commerce platform's actual inventory requires integrating outputs from a service providing popular product facets (features and their common values). This approach faces challenges: the facets service may not provide features for all product categories, or the range of values for a feature might be limited. Additionally, the LLM might sometimes struggle to generate relevant suggestions even when provided with features and values (e.g., suggesting a premium brand for a low-budget query). These challenges may be mitigated by improvements to the facets service and by fine-tuning smaller, specialized language models for this task.

\subsection{ArgsLLM}

ArgsLLM is a LLM-powered component responsible for extracting and resolving complete product names from the Standalone Query (SAQ) and conversation context. These extracted names are used by downstream pipelines to retrieve relevant product information and answer customer queries. ArgsLLM also handles customer queries that do not have a shopping intent, making it essential for both product-specific and general interactions.


\noindent This component is triggered when the detected coarse intent is one of the following, with ArgsLLM behaving as described:
\begin{itemize}[topsep=0pt,itemsep=1ex,partopsep=1ex,parsep=1ex]
    \item \textbf{\textit{answer\_product\_specific\_questions}:} ArgsLLM extracts the complete product name from SAQ, using the detected intent, the query itself, and a list of up to eight suggested products shown in the previous turn. Since customers rarely mention full product names, these suggestions are crucial for accurate extraction.
    \item \textbf{\textit{compare\_products}:} ArgsLLM identifies all product names referenced in the comparison query, again relying on the suggested product list to resolve incomplete names.
    \item \textbf{\textit{return\_direct\_response}:} ArgsLLM generates brief, courteous, and neutral replies for chitchat, or general knowledge queries, ensuring continued customer engagement.
\end{itemize}

\noindent If no products were suggested in the previous turn, or if the product of interest is missing from the suggested list, ArgsLLM extracts and returns the product name directly from the SAQ. If no product name is present in the query, it returns an empty response, which further triggers the follow-up module to prompt the user for the relevant product name. This design ensures robust handling of diverse user intents and query scenarios.

\subsection{Decision Assistant (DA)}

\begin{figure}[htbp]  
    \centering  \includegraphics[width=0.9\linewidth]{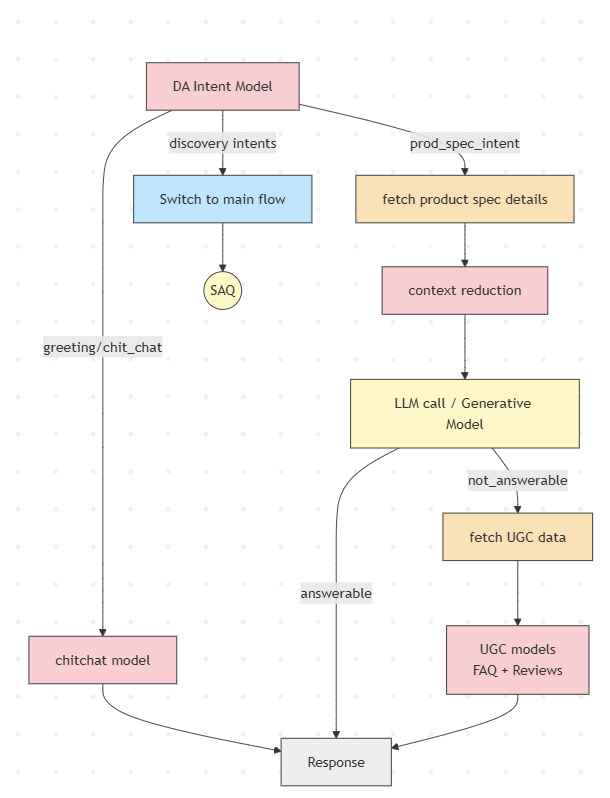}  
    \caption{DA architecture}  
    \label{fig:da-arch}  
\end{figure}

The main function of this flow is to answer user queries regarding a specific product. The queries can be on specifications, warranty, availability and other product related queries. In general these queries can be answered from either product catalog or user generated data like customer reviews, seller FAQs etc. Once the coarse intent model predicts the intent as DA, then the query is passed to ArgsLLM which gets the exact product name either based on the query and list of products returned from the latest search query in the session.
Given the query and product\_id, the user query is then passed to the DA intent model that decides if the query can be answered from the product catalog data, UGC data or other sources. Once the DA flow is triggered all the subsequent queries are directly routed to DA flow until the DA intent model predicts the query as discovery or other main intents.

In situations where the DA intent model interprets the user's intent as \textit{'answer\_product\_specific\_questions'}, the model will retrieve all the product specifications from the product catalog. For instance, if the product in question is a smartphone, the specifications might include the type of processor it uses (e.g., Snapdragon). From this pool of specifications, the top 20 most relevant specs are selected, based on their potential to answer the user's query. These chosen specifications, together with the user's query, serve as the context for a generative model (specifically and LLM) to formulate a fitting response to the user's query. If the generative model is unable to provide an answer, a template response is given, and a 'not\_answerable' flag is returned. This triggers the UGC flows, which involve selecting and presenting the top 3 FAQs and reviews that are most likely to address the user's query.

The selection of the most pertinent reviews and FAQs is facilitated by a bi-encoder-based semantic text similarity (STS) model. This model, trained using contrastive loss on proprietary e-commerce UGC FAQ data, is designed to assess the semantic similarity between texts. The same model, adapted for this specific domain, also functions as a context reducer, picking out the relevant specifications from the comprehensive list. This method of context reduction has proven to be nearly as effective as the fully supervised Dense Retriever (DPR) model. An added benefit is that it does not require retraining for different product categories, as the STS model is trained on data from a wide variety of categories.

\subsection{Summarization}

Summarization is essential for distilling complex product descriptions and user queries into concise, easily understood summaries, thereby enhancing user comprehension and decision-making. By focusing on relevant information, summarization reduces cognitive load and improves the overall user experience.

This component takes two inputs: the Standalone Query (SAQ) and detailed product information (including product name, features, and customer reviews). Using an LLM, we generate a summary tailored to the contextualized user query, ensuring coverage of product highlights and relevant pros and cons.

To ensure summaries are both concise and highly relevant to user queries, the summarization process employs targeted guidelines. Key product attributes are prioritized. A comprehensive summary is then constructed, addressing all pertinent objective and subjective facets of the product in relation to the query, while actively minimizing redundancy. This involves presenting positive or neutral aspects in a few descriptive sentences (e.g., up to three) to guide the user, alongside a brief mention of any relevant negative points (e.g., up to two) if identified in the source material. The overarching goal is to produce engaging, user-centric summaries that empower effective decision-making.

\subsection{Compare}
This flow is designed to assist users in comparing the features of two or more products. Given a compare query for example "which phone has a better camera?", we get the products retrieved by the latest search query in the current session and fetch their product details. We then apply context reduction to select top 20 most relevant feature required to address the query for each product and send this data to the LLM  prompting it to generate a comparison summary and final verdict for the query asked. The comparison summary here generates descriptive summary with the pros and cons of the feature(s) that the customer is interested in. The verdict is a single sentence based on the comparison summary to help the customer decide which product is better.

\subsection{CX/Post Purchase}
For post-purchase inquiries, the current system architecture primarily directs users to the e-commerce platform's established customer support channels. This typically involves guiding the customer to access their order history and initiate a support interaction regarding a specific purchase through the platform's existing help interfaces or dedicated support assistants.

\subsection{Other Flows}
In case user queries are not related to shopping context or if its regarding general FAQs on the e-commerce portal we have a set of flows which either provide pre-designed responses or answers these queries based on the FAQs.

\section{Evaluation Methods and Metrics}
One of the challenges inherent in conversational chat bot systems revolves around the evaluation of conversation quality with customers. In response to this challenge, we have devised a metric i.e. \textbf{Answerability}~\cite{gupta-etal-2022-answerability} which evaluates our assistant’s response quality at each turn of conversation and finally computes a metric based on the whole session.

For models which are part of the pipeline, specific metrics are allocated to comprehensively evaluate its performance over time. These metrics serve as benchmarks to assess the effectiveness and efficiency of each model's functioning throughout its operational duration within the pipeline.

\subsection{SAQ}
We track two metrics to evaluate the performance of SAQ, Query Modification Accuracy (QMA) and Product disambiguation Accuracy (PDA). The former metric deals with correctness of the generated query, in terms of retaining relevant features and not hallucinating unnecessary information. The second metric evaluates the correctness of the generated product name when user query warrants product disambiguation. 
In order to evaluate QMA, we follow two approaches:
\begin{itemize}
    \item LLM based - We developed a GPT-4 prompt to evaluate the accuracy of the reformulated query in terms of correct intent, budget, features and product category given the entire conversation history,
    \item Human based - We share the output of the top performing prompts on the evaluation dataset to human labelers to estimate the accuracy.
\end{itemize}
We also evaluate the accuracy of product disambiguation with the help of human labelers. 

\subsection{Coarse Intent Model}
 Since this is a classification model, we track traditional classification metrics like model accuracy, F1-score, Precision and Recall for evaluations.

\subsection{Summarization}

The summarization component of our chatbot is evaluated by assessing the quality of product summaries generated in response to user queries. To ensure a comprehensive assessment, we employ both human and tune LLM to match the perform of two main steps:

\begin{enumerate}[topsep=0pt,itemsep=-1ex,partopsep=1ex,parsep=1ex]
    \item \textbf{Product Relevancy:} The evaluator determines whether the product described in the summary matches the user's query. If the summary addresses the correct product, the ``Product Relevancy'' label is marked as True; otherwise, it is marked as False. For example, a summary about a refrigerator in response to a smartphone query would be marked as False.
    
    \item \textbf{Summary Quality:} The summary is further assessed on the following criteria:
    \begin{enumerate}[topsep=0pt,itemsep=-1ex,partopsep=1ex,parsep=1ex]
        \item \textit{Factual Accuracy:} The accuracy of product details in the summary is checked. If all facts are correct, the label is Pass; otherwise, it is Fail.
        \item \textit{Query Relevance:} The degree to which the summary addresses the user's query is rated as Fully Relevant, Partially Relevant, or Irrelevant. Partially Relevant and Irrelevant summaries are further analyzed to identify specific gaps.
    \end{enumerate}
\end{enumerate}

This structured approach enables us to systematically identify strengths and areas for improvement in the summarization component.

\subsection{Compare} 
The comparison evaluation is conducted using a large language model (GPT-4), which is provided with the user query, product context and specifications, the generated comparison summary, and a verdict. All key aspects are extracted from both the user query and the comparison summary. Each aspect is then evaluated according to two main criteria:

\begin{enumerate}[topsep=0pt,itemsep=-1ex,partopsep=1ex,parsep=1ex]
    \item \textbf{Relevancy:} Each aspect in the comparison summary is assessed for its connection to the user's query. An aspect is labeled as:
    \begin{itemize}[topsep=0pt,itemsep=-1ex,partopsep=1ex,parsep=1ex]
        \item \textit{Relevant} if it is directly mentioned in the query,
        \item \textit{Partially Relevant} if it can be reasonably inferred from the query, or
        \item \textit{Irrelevant} if it is not present or implied in the query.
    \end{itemize}
    \item \textbf{Correctness:} This criterion is divided into two parts:
    \begin{itemize}[topsep=0pt,itemsep=-1ex,partopsep=1ex,parsep=1ex]
        \item \textit{Verdict Correctness:} Evaluates whether the verdict accurately compares the specified aspect in the context of the user query. The verdict is labeled as \textit{Correct} if the comparison is accurate, \textit{Incorrect} if not, or \textit{N/A} if the aspect is not addressed.
        \item \textit{Comparison Correctness:} Assesses whether the comparison summary appropriately addresses or compares the specified aspect, considering the full context. This is labeled as either \textit{Correct} or \textit{Incorrect}.
    \end{itemize}
\end{enumerate}

This structured approach ensures a rigorous and transparent evaluation of both the relevance and accuracy of each aspect within the comparison component.

\subsection{ArgsLLM}
The ArgsLLM component was evaluated through a two-stage process involving both human and LLM-based annotations. Initially, human annotators and a larger LLM independently labeled model responses using the Standalone Query (SAQ), coarse intent, suggested product list, and generated output. Each response was marked as either \textit{good} or \textit{bad}, with corrections provided for \textit{bad} responses. Human annotators further categorized cases based on the presence and matching of product names in the query and content.

To scale and standardize evaluation, the evaluator LLM was subsequently tuned---leveraging advanced prompting strategies such as chain-of-thought reasoning---to closely match human annotation quality. Once aligned, the tuned evaluator LLM was systematically employed to assess model responses and provide detailed feedback, enabling efficient identification of failure points and targeted improvements for the ArgsLLM generation models.

For each coarse intent, clear labeling guidelines were followed. For \textit{answer\_product\_specific\_questions}, a response was labeled \textit{good} if it correctly handled product names as indicated by the query and context; otherwise, it was labeled \textit{bad}. For \textit{compare\_products}, responses were labeled \textit{good} if all relevant product names were accurately reflected, \textit{partially good} if only some were included, and \textit{bad} if key names were missing. For \textit{return\_direct\_response}, responses were labeled \textit{good} if they were succinct, polite, inoffensive, and neutral.

This iterative evaluation framework, combining human expertise and advanced LLM-based assessment, enabled scalable, high-quality evaluation and systematic enhancement of ArgsLLM’s generative capabilities.

\section{Experiments and Results}
\subsection{SAQ}

SAQ results obtained from LLM and human based evaluations are tabulated in Table \ref{tab:saq_1}. Our LLM based approach is strict and penalizes the module even for small mistakes. However, we are able to experiment and evaluate with a shorter turnaround time compared to human labeling. We chose the top performing prompts and perform human evaluation to obtain both QMA and PDA. The prompt with the highest QMA is chosen for production.

Table \ref{tab:turn} contains the turn wise accuracy of SAQ module. As the turn count increases, the accuracy of SAQ drops significantly. This is expected since the conversation gets more complex as the user converse for more turns. Only 449 sessions out of 4000 i.e $\approx 10\%$ of the sessions contain 5 turns. 

\begin{table}[h]
\centering
\begin{tabular}{|c|c|c|}
\hline 
 & \textbf{QMA} & \textbf{PDA} \\
\hline 
LLM based & 0.8949 & - \\
\hline 
Human based & 0.9234 & 0.9523 \\
\hline 
\end{tabular}
\caption{LLM and Human based evaluation}
\label{tab:saq_1}
\vspace{-10mm}
\end{table}

\begin{table}[h]
\centering
\begin{tabular}{|c|c|c|}
\hline 
\textbf{Turn} & \textbf{QMA} & \textbf{No. of datapoints} \\
\hline 
1 & 0.9865 & 4000 \\
\hline 
2 & 0.9057 & 1675 \\ 
\hline 
3 & 0.873 & 1063 \\ 
\hline 
4 & 0.8559 & 673 \\
\hline 
5 & 0.8396 & 449 \\
\hline 

\hline 
\end{tabular}
\caption{Turn wise accuracy of SAQ module with support}
\label{tab:turn}
\vspace{-9mm}
\end{table}

\subsection{Coarse Intent Model}

Table \ref{tab:coarse_intent_acc_1}  presents the accuracy metrics for various BERT base architectures tested on our dataset. We opted for the DistillBERT model due to its compatibility with our traffic and latency requirements. 

\begin{table}[h]  
\centering  
\begin{tabular}{|c|c|}  
\hline  
\textbf{Model} & \textbf{Test Accuracy} \\ \hline  
FK-Bert & 94.70\% \\  \hline 
FK-Bert + Mixup + Aug & 94.69\% \\  \hline 
BERT & 94.43\% \\  \hline 
DistillBert & 94.89\% \\  \hline 
DistillBert + Mixup + Aug & 94.75\% \\ \hline  
\end{tabular}  
\caption{Test accuracy of different models}  
\vspace{-6mm}
\label{tab:coarse_intent_acc_1}  
\end{table}  

Over time, we enhanced the model with the use of Active Learning and data adjustments. The final accuracy of the coarse intent model on our test dataset is approximately \textbf{97\%}. The accuracy details for the detailed intent model can be found in Table \ref{tab:coarse_intent_acc_2}.

\begin{table}[h]  
\centering  

\label{intent_accuracy}  
\begin{adjustbox}{max width=0.50\textwidth}  
\begin{tabular}{|l|c|c|c|}  
\hline  
\textbf{Intent} & \textbf{Precision} & \textbf{Recall} & \textbf{F1-Score} \\ \hline  
answer\_offer\_related\_questions & 0.9664 & 0.9839 & 0.9751 \\  
answer\_product\_specific\_questions & 0.4545 & 0.8333 & 0.5882 \\  
compare\_products & 0.5652 & 0.9455 & 0.7075 \\  
get\_answer\_from\_faq & 0.6875 & 0.7529 & 0.7187 \\  
not\_applicable & 0.7194 & 0.7542 & 0.7364 \\  
post\_purchase & 0.9895 & 0.9759 & 0.9827 \\  
return\_direct\_response & 0.8557 & 0.8706 & 0.8631 \\  
search\_for\_products & 0.9962 & 0.9867 & 0.9914 \\ 
\hline
accuracy & & & 0.9734 \\  
macro avg & 0.7793 & 0.8879 & 0.8204 \\  
weighted avg & 0.9771 & 0.9734 & 0.9748 \\ \hline  
\end{tabular}  
\end{adjustbox}  
\caption{Intent Accuracy}  
\vspace{-8mm}
\label{tab:coarse_intent_acc_2}  
\end{table}

\subsection{Summarization, Compare \& ArgsLLM}
Our summarization module demonstrated strong performance, achieving an overall factuality of \textbf{88.47\%} and a combined full and partial relevancy of \textbf{94.27\%} based on human annotations. Product relevancy for summarization was rated at \textbf{97.05\%} by human evaluators. 
For the compare module, GPT-4 evaluation yielded mean scores of \textbf{94.16\%} for relevancy, \textbf{89.44\%} for comparison correctness, and \textbf{88.18\%} for verdict correctness.
The ArgsLLM component was rated as \textit{good} in \textbf{90.24\%} of cases by human annotators, while GPT-4 evaluation reported a \textit{good} response rate of \textbf{92.33\%}.

\subsection{DA-QnA}

Table \ref{tab:Helpful_Answer_Rate} presents the accuracy of the DA question answering model across seven distinct Business Units. The results clearly shows that the model consistently maintains an average accuracy rate of approximately \textbf{90\%} or higher, demonstrating its effectiveness.

\begin{table}[h]  
\centering  
\begin{tabular}{|c|c|}  
\hline  
\textbf{BU/Category} & \textbf{Helpful Answer Rate} \\ \hline  
Mobiles & 89.65\% \\  \hline 
Clothing & 88.7\% \\  \hline 
Large & 94.2\% \\  \hline 
BGM & 90.2\% \\  \hline 
Home & 90.9\% \\  \hline 
Furniture & 88.1\% \\  \hline 
Electronics & 91.3\% \\ \hline  
\end{tabular}  
\caption{DA-QnA Metrics}  
\vspace{-8mm}
\label{tab:Helpful_Answer_Rate}  
\end{table}

\subsection{Answerability Evaluation}
Ground truth labels for the Answerability metric are obtained through human annotators and LLMs are also leveraged for assessing the bot's responses against user queries. Answerability is assessed at both turn and session levels, focusing on relevancy (Highly relevant, Partially relevant, Irrelevant) at the turn level, and overall successfullness (Successful, Unsuccessful) according to predefined labeling guidelines. Turnwise labels are used to calculate turn-level and session-level Answerability metrics. Session-level answerability demonstrated a consistent improvement of approximately 32\% over the evaluation period

\section{Impact}
The Flippi system was first deployed through an A/B experimentation framework. Since its initial deployment, the system has undergone numerous iterative development cycles, with each iteration yielding enhancements over its predecessor. During a major e-commerce sales event, Flippi was exposed to the entire user base through specific platform integrations, and concurrently A/B tested on primary user interfaces such as Search, Browse, and Category Landing Pages.
Following this period of broad exposure, the system's presence in certain platform-wide touchpoints has been adjusted, currently encompassing approximately 15\% of that initial scope, as part of ongoing strategic experimentation. The development of Flippi is an active process, characterized by continuous A/B testing and feature refinement to optimize user experience and system efficacy.

\begin{table}[h]
\centering
\begin{tabular}{|l|l|}
\hline
Engaged MAU           & 1.3M         \\ \hline
Last Touch Conversion & $\sim$1.0\%  \\ \hline
Thumbs-up share       & 68\%         \\ \hline
\end{tabular}
\caption{L0 Output Metricsand Thumbs-Up Share for Experience}
\label{tab:l0-output}
\vspace{-8mm}
\end{table}

Key Performance Indicators (KPIs) for Flippi include Engaged Monthly Active Users (MAU), Channel Conversion, and a user satisfaction and retention metric. An "Engaged MAU" is defined as a unique user who initiates at least one query with the assistant within a month. Table \ref{tab:l0-output} presents key output metrics, alongside a user satisfaction indicator.

\section{Conclusion and Future work}
Conversational assistants have fundamentally transformed the landscape of online shopping, and Flippi stands at the forefront of this revolution. By delivering a contextual, intuitive, and highly responsive shopping experience, Flippi empowers customers to make confident, informed purchase decisions with unprecedented ease. Its advanced natural language pipeline enables precise interpretation of customer queries, ensuring that users receive the most relevant and timely information available. Seamless integration with specialized assistants—including decision support, offer discovery, and expert help—further elevates Flippi into a truly end-to-end solution for digital commerce.

Flippi’s proven effectiveness is reflected in strong user engagement and satisfaction metrics, yet the journey toward excellence continues. Future enhancements will focus on refining individual system components to drive even greater accuracy in search relevance, intent prediction, and self-serve answer quality—key levers for raising our already impressive thumbs-up rate beyond 68\%. Exploring the deployment of in-house trained models promises to further strengthen Flippi’s conversational intelligence, enabling richer, more nuanced interactions that go beyond the capabilities of prompt-tuned commercial LLMs. Expanding language support to include regional and vernacular languages will unlock new opportunities for engagement and inclusivity, broadening Flippi’s reach across diverse customer segments.

Finally, a deeper investigation into Flippi’s long-term impact on customer loyalty and retention will yield valuable insights into the sustained value of conversational AI in e-commerce. As Flippi continues to evolve, it sets a new benchmark for customer-centric innovation, shaping the future of digital shopping and redefining what customers can expect from intelligent, conversational commerce.

\bibliographystyle{ACM-Reference-Format}
\bibliography{references}

\end{document}